\documentclass[sigconf]{acmart}

\AtBeginDocument{%
  }

\copyrightyear{2024}
\acmYear{2024}
\setcopyright{acmlicensed}
\acmConference[KDD '24]{Proceedings of the 30th ACM SIGKDD Conference on Knowledge Discovery and Data Mining}{August 25--29, 2024}{Barcelona, Spain}
\acmBooktitle{Proceedings of the 30th ACM SIGKDD Conference on Knowledge Discovery and Data Mining (KDD '24), August 25--29, 2024, Barcelona, Spain}
\acmDOI{10.1145/3637528.3671642}
\acmISBN{979-8-4007-0490-1/24/08}

\usepackage{multirow}
\usepackage{threeparttable}
\usepackage{enumitem}
\usepackage{diagbox}
\usepackage{xspace}
\usepackage{booktabs}

\newcommand{\vpara}[1]{\noindent\textbf{#1 }}
\newcommand{\equationref}[1]{Eq~\ref{#1}}
\newcommand{\secref}[1]{Section~\ref{#1}} %section reference
\newcommand{\figref}[1]{Figure~\ref{#1}} 
\newcommand{\tableref}[1]{Table~\ref{#1}} 
\newcommand{\modelname}{\textit{HomNet}\xspace}
 
\newcommand{\hide}[1]{} %hide

\begin{document}
\title{Chromosomal Structural Abnormality Diagnosis \\ by Homologous Similarity}
\author{Juren Li}
\authornote{Both authors contributed equally to this research.}
\email{jrlee@zju.edu.cn}
\orcid{0000-0001-8511-9921}
\affiliation{%
  \institution{Zhejiang University}
  \city{Hangzhou}
  \country{China}
}

\author{Fanzhe Fu}
\authornotemark[1]
\email{ffanz@zju.edu.cn}
\affiliation{%
  \institution{Zhejiang University}
  \city{Hangzhou}
  \country{China}
}

\author{Ran Wei}
\email{ranwei@diagens.com}
\affiliation{%
  \institution{Hangzhou Diagens Biotechnology Co., Ltd.}
  \city{Hangzhou}
  \country{China}
}

\author{Yifei Sun}
\email{yifeisun@zju.edu.cn}
\affiliation{%
  \institution{Zhejiang University}
  \city{Hangzhou}
  \country{China}
}

\author{Zeyu Lai}
\email{jerrylai@zju.edu.cn}
\affiliation{%
  \institution{Zhejiang University}
  \city{Hangzhou}
  \country{China}
}

\author{Ning Song}
\email{ningsong@diagens.com}
\affiliation{%
  \institution{Hangzhou Diagens Biotechnology Co., Ltd.}
  \city{Hangzhou}
  \country{China}
}

\author{Xin Chen}
\email{xin.21@intl.zju.edu.cn}
\affiliation{%
  \institution{Zhejiang University}
  \city{Hangzhou}
  \country{China}
  }

\author{Yang Yang}
\email{yangy@zju.edu.cn}
\authornote{Corresponding author.}
\affiliation{%
  \institution{Zhejiang University}
  \city{Hangzhou}
  \country{China}
}

\renewcommand{\shortauthors}{Juren Li et al.}
\begin{abstract}
Pathogenic chromosome abnormalities are very common among the general population. 
While numerical chromosome abnormalities can be quickly and precisely detected, structural chromosome abnormalities are far more complex and typically require considerable efforts by human experts for identification.
This paper focuses on investigating the modeling of chromosome features and the identification of chromosomes with structural abnormalities.
Most existing data-driven methods concentrate on a single chromosome and consider each chromosome independently, overlooking the crucial aspect of homologous chromosomes.
In normal cases, homologous chromosomes share identical structures, with the exception that one of them is abnormal.
Therefore, we propose an adaptive method to align homologous chromosomes and diagnose structural abnormalities through homologous similarity.
Inspired by the process of human expert diagnosis, we incorporate information from multiple pairs of homologous chromosomes simultaneously, aiming to reduce noise disturbance and improve prediction performance. 
Extensive experiments on real-world datasets validate the effectiveness of our model compared to baselines.
\end{abstract}




% make the title area
\maketitle

\section{Introduction}\label{sec:introduction}
chromosomal abnormalities, which include missing, extra, or irregular portions of chromosomal DNA~\cite{factsheet}, are the root cause of various genetic and hereditary diseases. 
It is estimated that 6 of every 1,000 newborns have chromosomal abnormalities, which often result in \textit{dysmorphism}, \textit{malformations}, and \textit{developmental disabilities}~\cite{shaffer2000molecular}.
These chromosomal changes caused by abnormalities lead to serious consequences, e.g. 25\% of all miscarriages and stillbirths, and 50\%–60\% of first-trimester miscarriages~\cite{hassold1980cytogenetic}.
With the help of chromosomal abnormality detection, clinicians can identify more abnormalities that may result in birth defects and take further preventive measures, which is of great significance to people’s livelihood and social development.

There are two types of chromosomal abnormalities: \emph{numerical abnormality} and \emph{structural abnormality}~\cite{queremel2023genetics}.
Numerical abnormalities involve the absence or addition of entire chromosomes ~\cite{gardner2012chromosome}. 
For example, an extra copy of chromosome 21 will cause \textit{Down syndrome}.
Structural abnormality refers to the deletion, excess, inversion, or translocation of a certain segment of one chromosome to another chromosome~\cite{alliance2009understanding}.
For example, a partial deletion of chromosome 4 causes \textit{Wolf–Hirschhorn syndrome}.
Numerical abnormalities can be easily detected by assessing the quantity of chromosomes.
However, structural abnormalities are more challenging as they are associated with chromosome mutation\footnote{Chromosome mutation means a change in a chromosomal segment, involving more than one gene~\cite{rieger2012glossary}.} ~\cite{VANCE2020525}, which makes it impossible to diagnose simply by the number of chromosomes as the numerical abnormality.
The detection of structural abnormalities demands the expertise of professionals well-versed in karyotype\footnote{
Karyotype is the overall visual representation of the entire set of chromosomes within the cells of an individual organism, including information about their sizes, quantities, and shapes~\cite{karyotype}.} analysis. 
Besides, to enhance the accuracy of diagnosing, clinicians typically need to analyze chromosomes from more than 10 cells for one patient. 
Given that there are 46 chromosomes in each human cell~\cite{gartler2006chromosome}, experts are required to visually inspect over 460 chromosome diagrams from the karyotypes.
Admittedly, this inspection process is time-consuming, labor-intensive, and therefore expensive.
Thus, we propose to design a deep learning-based model to automatically diagnose chromosomal structural abnormalities, aiming to expedite diagnosis, assist rapid decision-making, and reduce manual labor for clinicians, ultimately improving efficiency and accuracy in patient care.

\begin{figure}[t]
    \centering
    \includegraphics[width=0.9\linewidth]{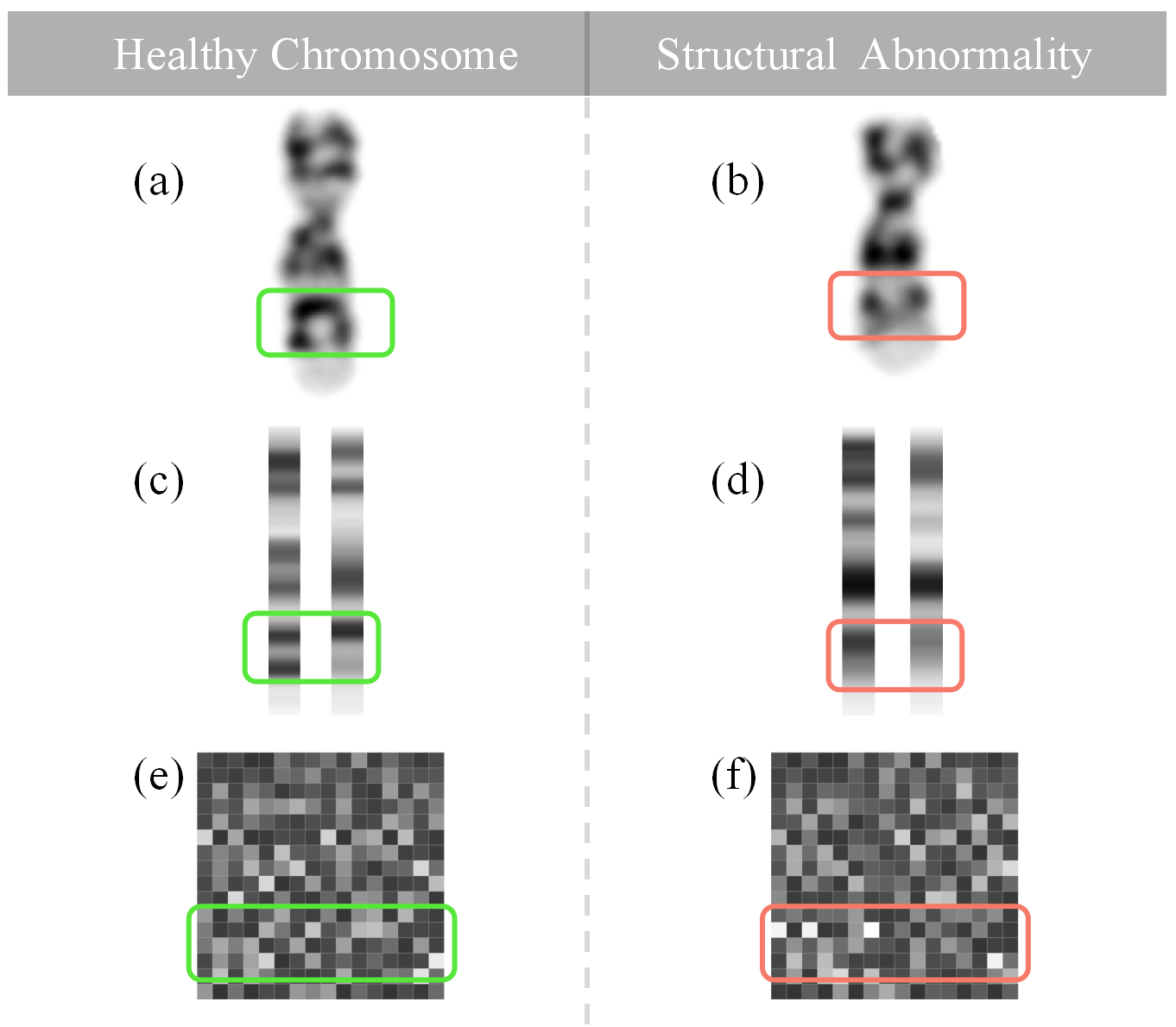}
    \caption{The homologous chromosomes with structural abnormality. (a) shows corresponding healthy region where structural abnormality occurred on (b). As for (c) and (d) are the data preprocessing we applied. (e) and (f) are the embedding of (c) and (d) by CMSBlock from HomNet model.}
    \label{fig:Sample_v1.1}
    \Description{A pair of homologous chromosomes with structural abnormality.}
\end{figure}

Despite the superior performance achieved by deep learning in various applications, challenges persist in detecting chromosomal structural abnormalities.
%data
The first challenge arises from the diverse morphology of chromosomes.
Structural abnormalities can occur on any of the 46 chromosomes, each with its own unique morphology, that is, different chromosomes may have different diagrams. 
Even within the same chromosome, variations exist due to their origins from different cells, potentially at different cell cycle~\cite{geiersbach2014subjectivity}, resulting in differing morphologies.
Thus, it is necessary to design of a method capable of detecting chromosomal structural abnormalities across various chromosomal morphologies.
The second challenge is the label imbalance. 
Compared to normal chromosomes, those with structural abnormalities are relatively rare.
Indeed, in our data, less than 4\% of the chromosomes have structural abnormalities. 
While the imbalance problem has been addressed in many fields~\cite{johnson2019survey}, to the best of our knowledge, it has not been well studied in the context of detecting chromosomal structural abnormalities.
The third challenge lies in the inconsistent data distribution.
Chromosomes obtained from different hospitals may exhibit variations in their chromosomal diagrams due to differences in equipment and environmental conditions. 
Additionally, the distribution of type of structural abnormalities varies from hospital to hospital, resulting in a different distribution of data from different hospitals.
These factors make the data distribution of chromosomes inconsistent and make it challenging to diagnose the structural abnormalities.

To address the aforementioned challenges, we propose a domain knowledge based framework, \modelname, to model chromosomes and then diagnose structural abnormalities.
To tackle the first challenge, we aim to capture common characteristics of structural abnormalities across chromosomes with varying morphologies. 
For a pair of homologous chromosomes\footnote{Homologous chromosomes refer to a pair of chromosomes in a biological organism's cell that share similar gene sequences and structural features. Typically, one chromosome in the pair is inherited from the father, while the other is inherited from the mother.~\cite{carroll2015introduction}}, their diagrams exhibit high similarity, since they share the same chromosome type and in the same cell.
However, when a pair of homologous chromosomes has a structural abnormality, their diagrams typically manifest differences in the region where the abnormality occurs, as illustrated in ~\figref{fig:Sample_v1.1}.
Therefore, we propose to diagnose structural abnormalities by capturing the differences between homologous chromosomes based on the homologous similarity.
% challenge 2
Addressing the label imbalance problem, where there are only a few abnormal samples, we utilize a large number of normal chromosomes from healthy individuals to construct artificial structural abnormalities based on the occurrence of real-world structural abnormalities. 
Then, a self-supervised task is designed to train the \modelname to detect pairs of homologous chromosomes with artificial structural abnormalities, enabling \modelname to diagnose structural anomalies using homology similarity.
% challenge 3
To mitigate the problem of inconsistent data distribution across different hospitals, we employ pre-training strategy with self-supervised tasks to pre-train \modelname and then fine-tune it on different hospital datasets. 
Fine-tuning enables the model to adapt to different data distributions.
Thus, the problem of inconsistent distribution can be overcome.
In addition, we collected data of 61350 chromosomes diagrams from 102 real-world patients with structural abnormalities and data of 458876 chromosomes diagrams from 1581 real-world healthy people.
The data is anonymized to protect the privacy. 
The ample number of chromosome data from healthy people enables \modelname to capture common characteristics of chromosomes and develop the capability to diagnose structural abnormalities.

\modelname integrates an end-to-end pipeline with AutoVision~\cite{song2021chromosome}, an AI analysis system for karyotypes.
And it has undergone multicenter clinical trials in three Grade A tertiary hospitals in China.
Overall, the contribution of this article can be summarized as follows:

\begin{itemize}[leftmargin=*]

\item To the best of our knowledge, we are the first to study data-driven methods for identifying chromosomal structural abnormalities. 
\item Inspired by domain knowledge, we introduce \modelname, designed to diagnose chromosomal structural abnormalities through homologous similarity. 

\item We propose a self-supervised learning strategy to train \modelname, enabling it to capture chromosome features and address the challenge of inconsistent distribution across different hospitals.
\item We conduct extensive experiments on real-world dataset involving multiple chromosomes with structural abnormalities. 
Our results validate the effectiveness of \modelname on chromosomal structural abnormality diagnosis. 

\end{itemize}

\section{Preliminaries}

To clearly illustrate our work, this section first introduces the dataset and the preprocessing steps. 
Detailed and formal definitions of the dataset and the problem formulation will be provided in this section.

\subsection{Dataset and Preprocessing}
\label{text:Dataset}
Each chromosome data in the dataset contains two sequences sampled from the chromosome diagram.
As shown in \figref{fig:Sample_v1.1} (a) and (b), the chromosome diagrams indicate a unidirectional flow of information, suggesting that the information density aligns with the chromosome’s direction.
Typically, image data exhibits information density along two perpendicular directions. 
Therefore, we propose converting the original data into chromosomal sequences for our dataset.

\begin{figure}[t]
    \centering
    \includegraphics[width=0.3\linewidth]{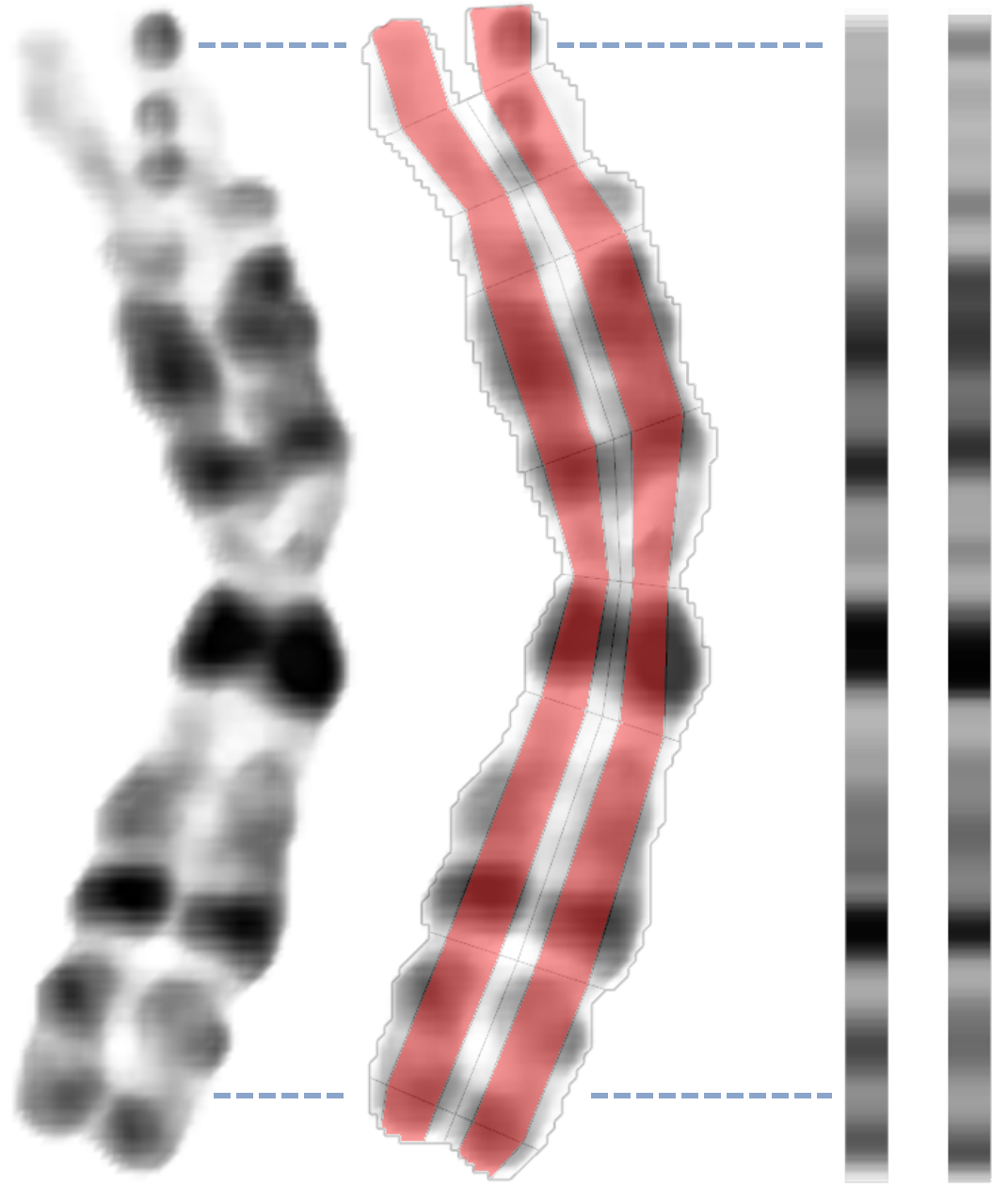}
    \caption{The chromosomal sequences are the gray mean sequences of the left and right parts of the chromosome diagram.}
    \label{fig:Sample_v2.1}
    \Description{How chromosome sequence data is obtained.}
\end{figure}

As shown in Figure~\ref{fig:Sample_v2.1}, the chromosomal sequences represent the gray mean sequences derived from the left and right parts of the corresponding chromosome diagram.
In a chromosome diagram, sister chromatids cross over the left and right sides of the chromosome. 
A sister chromatid refers to the identical copies (chromatids) formed by the DNA replication of a chromosome~\cite{peters2012sister}.
In theory, both of them would look the same in chromosome diagram. 
However, the left and right sides of the chromosome diagram are not completely symmetrical due to the bending of the chromosome and uneven staining of stains. 
Thus, we need to sample both the left and right sequences. 
For convenience, we use the term``chromosome'' to represent the sampled chromosomal sequence.

In addition to data from diagrams, the corresponding chromosome type and band level\footnote{
The bands on chromosomes correspond to distinct segments of the DNA molecule, and the band level is determined by the number of bands in a set of distinct chromosomes~\cite{geiersbach2014subjectivity}.} 
can be obtained. 
Human beings have 22 pairs of autosomes (common to both men and women) and 1 pair of sex chromosomes (XX for women, XY for men), resulting in 24 types of chromosomes. 
Different cells may be in different cell cycles ~\cite{abercrombie2017dictionary}, which makes their chromosomes different in shape. 
In other words, the same chromosomes in different cells may have different band levels. 
We collected chromosomal data from four hospitals, denoted as $\text{Hos\#1}$, $\text{Hos\#2}$, $\text{Hos\#3}$ and $\text{Hos\#4}$ to meet privacy and data security requirements. 
The respective patient counts for these hospitals are 24, 27, 30, and 21. 
For each patient, we gathered multiple chromosomes from different cells, including both normal and abnormal chromosomes. 
Each patient exhibited only one type of chromosomal structural abnormality. 
After excluding unidentifiable chromosomes due to mutual occlusion between chromosomes and excessive bending, each hospital’s dataset contained 14196, 14840, 16920, and 13496 pairs of normal homologous chromosomes, along with 428, 457, 590, and 378 pairs of homologous chromosomes with structural abnormalities. 
The proportion of abnormal chromosomes in each dataset was less than 4\%. 
Additionally, we also collected chromosomal data from 1581 healthy individuals at $\text{Hos\#1}$, resulting in a total of 458876 pairs of homologous chromosomes.

The subsequent table provides breakdown of patient samples across each hospital:
\begin{table}[htbp]
\centering
\caption{Hospital Sample Data}
\label{tab:hospital_samples}
\begin{tabular}{|c|c|c|c|}
\hline
\textbf{Hospital} & \textbf{Patient} & \textbf{Normal sample} & \textbf{Abnormal sample} \\ \hline
Hos\#1 & 24 & 14196 & 428 \\ \hline
Hos\#2 & 27 & 14840 & 457 \\ \hline
Hos\#3 & 30 & 16920 & 590 \\ \hline
Hos\#4 & 21 & 13496 & 378 \\ \hline
\end{tabular}
\end{table}

\subsection{Problem definition}

Inspired by the diagnostic process of human experts, we find that considering the information of multiple-pairs homologous chromosomes at the same time can make the judgment more reliable. This also reduces the noise disturbance of considering only one pair of homologous chromosomes. Thus we combine multi-pair homologous chromosomes to form a \emph{bag}, denoted by $X$.

Each homologous chromosome in $X$ is denoted by $H_i$.

Then we get $X=\{H_1, H_2, \dots, H_m\}$, where $m$ is the number of homologous pairs.
All the homologous chromosomes $H_i \in X$ are from different cells of the same patient, and their chromosome type are the same, denoted by $c$. 
$c$ is one-hot vector with the length of 24. 
As mentioned above, the band levels of chromosomes of cells may be different. 
In the dataset, band level is denoted by $b$, a one-hot vector with the length of 4. 
Because there are four types of band level in our dataset: 300, 400, 550 and 700, which also covers most of the common band types.
Each $H_i$ consists of two chromosomes, where homologous chromosome 1 and 2 are denoted by $H_{i, 1} \in \mathbb{R}^{2\times d}$ and $H_{i, 2} \in \mathbb{R}^{2\times d}$, respectively. $d$ is the length.

The chromosomal structural abnormalities detection task is: given a query tuple $(X, c, b)$ containing the chromosomes bag $X$ and the corresponding information (chromosome type $c$ and band level $b$), determining whether this query tuple $(X, c, b)$ is from chromosomes with structural abnormalities.

\section{Methodology}
\label{text:Methodology}
In this section, we introduce the proposed model \modelname for diagnosing chromosomal structural abnormalities by homologous similarity.

\begin{figure*}[ht]
    \centering
    \includegraphics[width=1.0\textwidth]{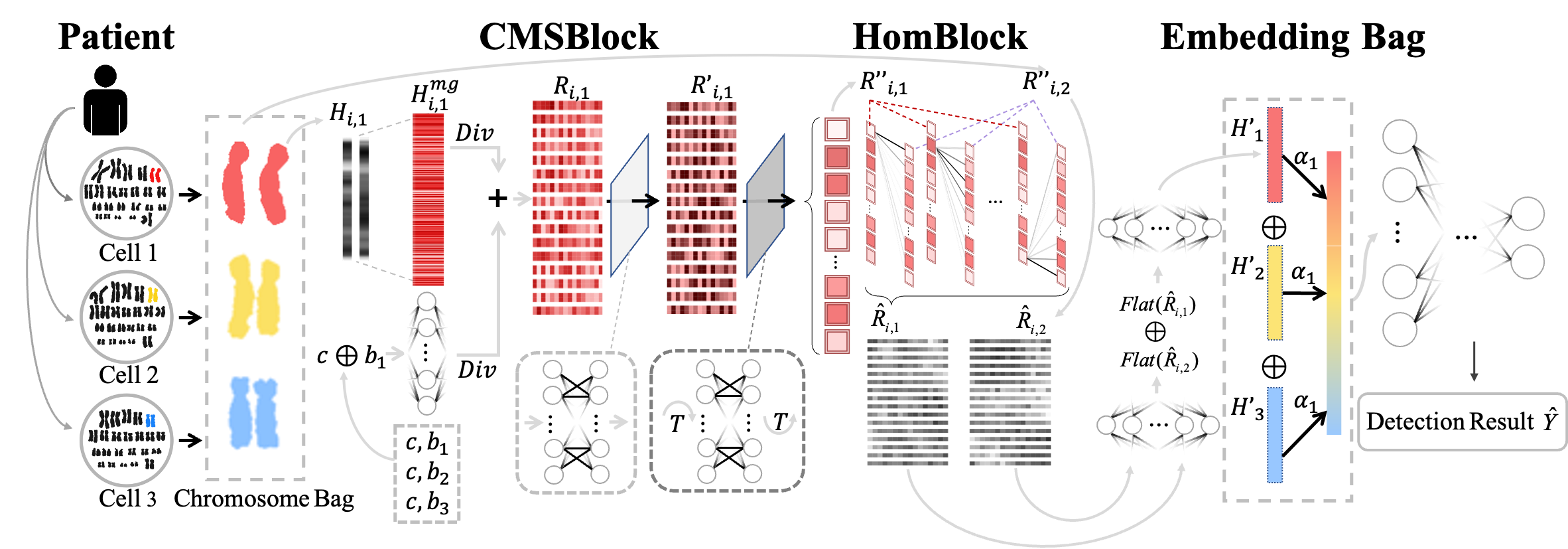}
    \caption{The framework of \modelname. The arrow heads refer to the dataflow of a pair of homologous chromosomes.}
    \label{fig:HomNet}
    \Description{The frame work of \modelname}
\end{figure*}
 
\subsection{Overview}
The framework of \modelname is depicted in \figref{fig:HomNet}, comprising three main components: $CMSBlock$, $HomBlock$, and $BagBlock$. 
 
To capture common characteristics of structural abnormalities across chromosomes with varying morphologies, we design \modelname based on homologous similarity.
We generate five types of artificial abnormal chromosome sequences as negative samples based on real-world structural abnormalities.
Subsequently, we devise a self-supervised training strategy to utilize the chromosomes of healthy individuals and the artificially generated chromosomes.
To address the data inconsistent distribution problem across different hospitals, we propose fine-tuning the model for each specific hospital.

For each query tuple $(X, c, b)$, $CMSBlock$ extracts the distinguishable features $R_{i, j}^{\prime \prime}$ from each chromosome in $X$, where $j\in{1, 2}$. 
Subsequently, $HomBlock$ aligns each pair of homologous chromosomes and captures the differences between them based on homologous similarity, denoted as $H_{i}^{\prime}$.
Finally, $BagBlock$ considers all the $H_{i}^{\prime}$ to diagnose the structural abnormalities. 

More specifically, $CMSBlock$ first captures the features of the chromosomes.
During this process, $CMSBlock$ utilizes prior information band level $b$ and chromosome type $c$ to help \modelname to extract the features.
Generally, there is at most one chromosome with structural abnormality in any pair of homologous chromosomes.
Thus, we propose to detect the chromosomal structural abnormalities by homologous similarity. 
However, it is challenging to capture the differences since the shape of each chromosome varies and the diagram of chromosome pair may be flipped over, \emph{i.e.}, they are not perfectly aligned.
To align each part, $HomBlock$ uses an inter-attention based strategy to capture the difference $H_{i}^{\prime}$ between the representations $H_{i, 1}^{\prime}$ and $H_{i, 2}^{\prime}$ of homologous chromosome 1 and 2. 
As mentioned above, in order to reduce noise disturbance and increase judgment reliability, \modelname considers multiple pairs of homologous chromosomes simultaneously in $BagBlock$. 
After getting all the difference representation $(H_{1}^{\prime}, H_{2}^{\prime}, \dots, H_{m}^{\prime})$ of the Bag $X$, $BagBlock$ makes a prediction $\hat{Y}$ according to these representations.

\subsection{Self-supervised Learning with Artificial Abnormalities}
\label{text:BagAndSSL}
Due to the significant imbalance between the number of chromosomes with structural abnormalities and those without, it becomes challenging to identify distinctive features for abnormality detection~\cite{japkowicz2002class}. 
On the bright side, we possess a substantial amount of chromosome data from healthy individuals. 
Leveraging this data alongside artificially generated structural abnormalities, we pretrain the model. 
To simulate various types of structural abnormalities observed in real-world scenarios, we create five types of artificial chromosomes with structural abnormalities, as detailed in the Appendix. 
Our self-supervised learning task involves classifying normal and artificial abnormal chromosome pairs to capture diverse structural abnormality manifestations.

To prevent \modelname from solely relying on the presence of artificial data when assessing abnormal chromosome pairs, we also incorporate chromosome pairs from various types as abnormal samples.
Given the length discrepancies among different chromosome types, we categorize the chromosomes into seven groups based on their length~\cite{mcgowan2020iscn}, ensuring each group of chromosomes is similar in length.
Consequently, we exclusively form abnormal chromosome pairs using chromosomes from the same group within a cell.  
This approach enables our model to capture chromosome features and understand fundamental structural abnormality characteristics.
Subsequently, we fine-tune the model using chromosomes from different hospitals.

%%%%%%%%% CMSBlcok
\subsection{Modeling the Feature of Chromosomes}
Before capturing the differences between homologous chromosomes, we model features of chromosomes. 
Each chromosome, $H_{i, j}$, is from a chromosome diagram with or without structural abnormality. 
% Therefore, we can extract the chromosome features according to the corresponding sequence data. 
% Then, we compare the differences between the homologous chromosomes according to the extracted features. 
We propose $CMSBlock$ to extract the features of each chromosome. 
As \figref{fig:HomNet} shows, the $CMS Block$ first merges two sequences into one, and get $H_{i, j}^{mg}$. 
Then, $CMSBlock$ utilizes $H_{i, j}^{mg}$, band type $b$ and chromosome type $c$ to obtain the chromosomal representation $R_{i,j}^{\prime\prime}$.

\vpara{Merge two chromosome sequences.}
The chromosome data we obtain consists of two sequences: $H_{i, j} \in \mathbb{R}^{2\times d}$. 
As previously mentioned, these sequences represent the mean sequences of the left and right portions of the chromosome diagrams, respectively. 
Given the partial differences between the left and right sequences within $H_{i, j}$, both containing crucial chromosome information, a simple merging approach using average, maximum, or minimum values would result in substantial information loss. 
Therefore, we opt to merge these sequences using a learnable method. 
Specifically, we employ a convolution operation, and get the representation $H_{i, j}^{mg}$.
\begin{equation}
    H_{i, j}^{mg} = Conv(H_{i, j})
    \label{eq:mg}
\end{equation}

\vpara{Model the feature of regions.}
Subsequently, $CMSBlock$ extracts the features of each chromosome, obtaining the feature representation $R_{i, j}^{\prime \prime}$. 
As depicted in \figref{fig:Sample_v2.1}, the chromosome diagram displays multiple grayscale bands along the chromosomes, which are referred to as regions~\cite{mcgowan2020iscn}. 
When a structural abnormality occurs, the region where the abnormality happens will correspondingly change.
For instance, in the case of a chromosomal deletion~\cite{mcgowan2020iscn}, the corresponding region will also be missing.

Therefore, \modelname detects structural abnormalities by analyzing the differences in regions between homologous chromosomes.

Therefore, we divide $H_{i, j}^{mg}$ into multiple subsequences, each representing a region of the chromosome. 

Additionally, leveraging the prior knowledge that chromosome types and band levels can aid in modeling the chromosomes, we fuse the feature of chromosome type $c$ and band level $b_i$ into each subsequence.
\begin{equation}
    R_{i, j} = Div(W_{info}(c \oplus b_i)) + Div(H_{i, j}^{mg})
    \label{eq:cmsblock1}
\end{equation}
where $\oplus$ denotes the concatenation operation and $Div$ represents the segmentation operation. 

For $H_{i, j}^{mg}$, the $Div$ operation is carried out as per \equationref{eq:mg}, using convolution with a kernel size of $(2, k^{mg})$, a stride of $(2, k^{mg})$, and with $l_r$ output channels. 
Consequently, the output after the convolution produces a matrix where each row, having a length of $l_r$, corresponds to a region of length $k^{mg}$.
As for $W_{info}(c \oplus b_i)$, the $Div$ operation is implemented through reshaping, where $W_{info}$ denotes a learnable parameter matrix. 
Thus, the matrix of regions representation $R_{i, j} \in \mathbb{R}^{n_r \times l_r}$, where $n_r$ represents the number of regions depending on $k^{mg}$ and the length of $H_{i, j}$. 

For each region representation in $R_{i, j}$, we use the same method to extract similar texture features, which is also called weight sharing:
\begin{equation}
    R_{i, j}^{\prime} = \sigma(R_{i, j}W_{R1})W_{R2}+R_{i, j}
    \label{eq:cmsblock2}
\end{equation}
where $W_{R1}$ and $W_{R2}$ are learnable parameter matrices. 
\equationref{eq:cmsblock2} extracts the features of each chromosome region, also referred to as local features.

However, there are also some types of structural abnormalities where a region of a chromosome can be found in a similar region of another normal chromosome. 
For example, in chromosome duplication abnormality~\cite{li2005expression}, the duplicated region originates from a part of the normal chromosomes. 
Consequently, \modelname may struggle to capture the differences between homologous chromosomes, as their corresponding regions are similar. 
Despite this, a particular structural abnormality, the Robertsonian translocation, indicates that an entire chromosome has attached to another ~\cite{therman1989nonrandom}. 
This results in data with this structural abnormality exhibiting more texture features due to the addition of another chromosome. 
Thus, capturing the global features of the chromosome will help the model better detect such abnormalities. 
By transposing $R_{i, j}^{\prime}$, $CMSBlock$ can capture the cross-region features, also referred to as global features.
\begin{equation}
    R_{i, j}^{\prime \prime} = (\sigma({R_{i, j}^{\prime}}^{T}W_{R3})W_{R4})^{T}+R_{i, j}^{\prime}
    \label{eq:cmsblock3}
\end{equation}
After feature extraction, we obtain the segmented features representation of chromosomes, $R_{i, j}^{\prime \prime}$. 

%%%%%%%%%%HomBlock
\subsection{Capture the Differences between Homologous Chromosomes}
Based on expert detection processes, \modelname diagnoses structural abnormalities through homologous similarity.
As depicted in ~\figref{fig:Sample_v1.1}, diagrams of homologous chromosome pairs with structural abnormalities differ, while those of normal homologous chromosomes are similar.
However, most existing anomaly detection methods concentrate on individual objects and do not leverage the distinctions between homologous chromosomes with structural abnormalities. 
Therefore, we introduce $HomBlock$ to capture the differences between homologous chromosomes, based on their feature representations, $R_{i, 1}^{\prime \prime}$ and $R_{i, 2}^{\prime \prime}$. 

\vpara{Adaptive region alignment in homologous pairs.}
As mentioned above, obtaining the difference of homologous chromosomes involves comparing the distinctions between the feature representations of corresponding regions in a pair of homologous chromosomes. 
However, due to the free oscillation of chromosomes in cells, and potential inversion of chromosome diagrams, the representations $R_{i, 1}^{\prime \prime}$ and $R_{i, 2}^{\prime \prime}$, which encode chromosomal structural information, are not well aligned. 
Mere subtraction or stacking of these representations without proper alignment introduces bias. 
Therefore, $HomBlock$ employs an adaptive alignment method to capture homologous differences by aligning each region between homologous chromosomes. 
$HomBlock$ adaptively aligns each row of $R_{i, 1}^{\prime \prime}$ with each row of $R_{i, 2}^{\prime \prime}$. 
Here, $r_{i, j}^e$ represents the $e_{th}$ row of $R_{i, j}^{\prime \prime}$ and also the $e_{th}$ region in the chromosome.

$HomBlock$ initially computes the similarity between corresponding regions.
For instance, considering the $e_{th}$ region of homologous chromosome 1, $r_{i, 1}^{e}$, we calculate the similarity between all regions of homologous chromosome 2.  
Subsequently, we utilize the similarity score as the alignment coefficient, allowing $r_{i, 1}^{e}$ to focus more on regions with high similarity. 
This process can be achieved through an attention mechanism.
\begin{equation}
    a_{i, 1}^{e, z}=(W_q r_{i, 1}^{e})\cdot(W_k r_{i, 2}^z)/\sqrt{d_a}
    \label{eq:HomBlock1}
\end{equation}

\vpara{Capture the differences in each region.}
~\equationref{eq:HomBlock1} denotes the similarity between the $e_{th}$ region of homologous chromosome 1 and the $z_{th}$ region of homologous chromosome 2.
Here, $d_a$ denotes the dimension of $(W_q r_{i, 1}^{e})$ and $(W_k r_{i, 2}^{z})$. 
Once the similarity score is obtained, $HomBlock$ adaptively aligns the regions based on the similarity score. 
This alignment process allows the capture of differences between regions, providing crucial information for \modelname to detect structural abnormalities.
\begin{equation}
    \tilde{r}_{i, 1}^{e} = \frac{\sum_{z=1}^{n_r}a_{i, 1}^{e, z}(W_v r_{i, 2}^{z})}{\sum_{z=1}^{n_r}a_{i, 1}^{e, z}}
    \label{eq:HomBlock2}
\end{equation}
$\tilde{r}_{i, 1}^{e}$ represents the aligned result of the $e_{th}$ region in homologous chromosome 1. 
Subsequently, $HomBlock$ captures the difference between $r_{i, 1}^{e}$ and the aligned result $\tilde{r}_{i, 1}^{e}$.  
In our approach, we employ a multi-headed attention mechanism for alignment
Specifically, $\tilde{r}_{i, 1}^{e, 1}$ and $\tilde{r}_{i, 1}^{e, 2}$ denote the alignment results from the first and second attention heads, respectively. 
Thus, the differences between extracted regions can be expressed as:
\begin{equation}
    \hat{r}_{i, 1}^{e} = W_{diff}(r_{i, 1}^{e}+W_{head}(\tilde{r}_{i, 1}^{e, 1}\oplus \tilde{r}_{i, 1}^{e, 2} \oplus \dots \oplus \tilde{r}_{i, 1}^{e, n_h}))
    \label{eq:HomBlock3}
\end{equation}
where $n_h$ is the number of heads. 
$\hat{r}_{i, 1}^{e}$ represents the difference feature representation of the $e_{th}$ region. 
The collection of all obtained $\hat{r}_{i, j}^{e}$ forms the difference matrices $\hat{R}_{i, j}$. 
Consequently, we integrate these difference features of the regions to obtain the difference representation $H_i^{\prime}$ of the homologous chromosome pair:
\begin{equation}
    H_{i}^{\prime} = \sigma(W_{hom}(Flat(\hat{R}_{i, 1})\oplus Flat(\hat{R}_{i, 2})))
    \label{eq:HomBlock4}
\end{equation}
where $Flat()$ is to flatten matrix to vector and $W_{hom}$ is a learnable parameter matrix.

\vpara{Resolve data distribution inconsistency by multi-pair homologous chromosomes.}
As discussed in ~\secref{text:Dataset}, the data may be susceptible to noise disturbance. 
Relying solely on one pair of homologous chromosomes could potentially lead to less accurate predictions.
Hence, we introduced $BagBlock$ to enable a comprehensive prediction by incorporating data from multiple pairs of homologous chromosomes. 
This approach draws inspiration from the diagnostic process of human experts. 
Similar to how human experts diagnose chromosomal structural abnormalities by considering multiple chromosome diagrams, our method aims to improve the reliability of diagnostic results. 
By simultaneously integrating information from multiple pairs of homologous chromosomes, \modelname not only enhances the reliability of detection results but also mitigates the impact of noise disturbance.
Structural abnormalities in some chromosome diagrams are obvious, while others may be obscure. 
This is why experts need to combine chromosome information from multiple cells to make a diagnosis.  
In this case, the contribution of the difference representation $H_{i}^{prime}$ to the identification of the structural abnormality varies.  
Simply adding or concatenating them directly may not yield optimal results. 
Therefore, we compute an aggregate weight $\alpha_{i}$ for the abnormal representation $H_{i}^{prime}$. 
The abnormal representations $H_{i}^{\prime}$ are then summed based on the aggregate weight $\alpha_{i}$, and \modelname makes predictions.
\begin{equation}
    \alpha_{i} = Sigmoid(MLP(H_{i}^{\prime})) 
\end{equation}
\begin{equation}
    \hat{Y} = \sigma(W_{bag}(\Sigma_{i=1}^{m}\alpha H_{i}^{\prime}))
\end{equation}
where $W_{bag}$ are learnable parameter matrices and $\hat{Y}$ is the prediction result. 
Ultimately, we employ cross-entropy loss as the objective function to optimize \modelname.
\begin{equation}
    \mathcal{L} = L_{CE}(\hat{Y}, Y)
\end{equation}
where $Y$ is the ground truth.

\section{Experiment}
\subsection{Experiment Setting}
\vpara{Datasets}. 
As mentioned in \secref{text:Dataset}, we use the dataset sampled from healthy people for self-supervised learning. 
Additionally, we also use the artificial samples with and without structural abnormalities. 
After self-supervised training, we fine tune \modelname on each dataset from patients with chromosomal structural abnormalities from $Hos\#1$, $Hos\#2$, $Hos\#3$ and  $Hos\#4$. 
The number of homologous chromosome pair in a bag is set as 5, that is, $m$ is set as 5.
Due to the privacy and sensitivity of chromosome data with structural abnormalities, almost all public datasets contain only normal chromosomes.
Therefore, we constructed artificial structural abnormal chromosome data on the public chromosome dataset Pki-3~\cite{906069}.
Experiments are also performed on this dataset for supplemental reference.
More details about the dataset can be found in the Appendix.

\vpara{Baseline.}  
We compare our proposed model, \modelname, with several baseline models.
Specifically, we initially compare our model with a classical machine learning method: Logistic Regression (LR) based on feature engineering. 
Additionally, we conduct experiments using commonly employed unsupervised methods for anomaly detection, including OC-SVM~\cite{scholkopf2001estimating}, Deep-SVDD~\cite{ruff2018deep}, and an AutoEncoder (AE) based method~\cite{zhao2019pyod}. 
Furthermore, we compare our model with supervised methods, such as MLP-Mixer~\cite{tolstikhin2021mlp}, Vit~\cite{dosovitskiy2020image}, Xception~\cite{chollet2017xception}, ResNet~\cite{he2016deep}, DensNet~\cite{huang2017densely}, and MobileNet~\cite{howard2017mobilenets}. 
In addition, we also conduct an experiment with a recent contrastive learning-based method for detecting structural anomalies~\cite{bechar2023highly}. 
This method employs a Siamese architecture and allows for interchangeable backbone models, hence it is abbreviated as ``Siamese''.

\vpara{Training.}
For the pre-training stage, we utilize normal samples and samples with artificial abnormalities. During this stage, we implement Adam as the optimizer, setting learning rate at 0.001 with a batch size of 512. A validation set, which does not overlap with the training set, is also used to identify the optimal pre-trained model. We then incorporate an early stopping mechanism to monitor model convergence closely. Specifically, if there's no observed enhancements in the validation set over 10 consecutive epochs, the training is discontinued. For the fine-tuning stage, we freeze the parameters of CMSBlock and the first layer of HomBlock. This strategy helps transfer learned knowledge from the pre-training stage. The Adam optimizer, with a learning rate of 0.00001, is once again used here. We also employ the same early stopping mechanism here.

\vpara{Evaluation metrics and implementation details.} 
We evaluate the prediction preformance of \modelname and baseline methods with F1 score and AUC-ROC score,  capturing both the precision-recall balance and the model’s overall ranking ability across varying decision thresholds.
Besides, we use Adam~\cite{kingma2014adam} optimizer with a learning rate of 0.001 for self-supervised learning and that of 0.00001 for fine-tuning, and the batch size is set as 512. 
During the fine-tuning, we freeze the parameters of $CMSBlock$ and the first difference extraction layer in $HomBlock$, namely the first attention layer. 
More details on the baseline and experimental setup can be found in the Appendix.
The code is available at \url{https://github.com/zjunet/HomNet}.

\begin{table}[htbp]
    \centering
    \caption{Experimental results on four real-world hospital datasets.}
    \resizebox{1.0\columnwidth}{!}{
    \setlength{\tabcolsep}{0.8pt}
    \begin{threeparttable}
        \begin{tabular}{c c c c c c c c c}
        \toprule
        % first line
        \multirow{2}{*}{\diagbox{Method}{Hospital}} & 
        \multicolumn{2}{ c}{$Hos\#1$}& 
        \multicolumn{2}{ c}{$Hos\#2$}& 
        \multicolumn{2}{ c}{$Hos\#3$}& 
        \multicolumn{2}{ c}{$Hos\#4$}\\
        % hos_1 for ZD3, hos_2 for ZD1, hos_3 for JH, hos_4 for NB
        % second line
        & AUC& F1
        & AUC& F1
        & AUC& F1
        & AUC& F1 \\
        % third line
        \midrule
        $LR$ & 
        % hos_1 for ZD3, hos_2 for ZD1, hos_3 for JH, hos_4 for NB
        % AUC   F1  |  AUC     F1 |   AUC   F1  |   AUC  F1
        72.43& 19.51& 75.23& 20.54& 81.72& 23.47& 75.56& 19.10\\
        % \midrule
        OC-SVM& 49.8& 11.0& 47.9& 9.23& 65.6& 13.97& 17.38& 17.38\\
        % \midrule
        $Deep-SVDD$ & 
        46.91& 10.28& 59.31& 12.56& 64.83& 14.67& 51.13& 13.01\\
        % \midrule
        $AE$& 
        60.27& 14.78& 63.74& 21.14& 65.82& 17.63& 61.41& 21.50\\
        \midrule
        %%%%%%%supervised
        $MLP-Mixer$ & 
        90.76& 50.02& 92.83& 56.27& 85.24& 39.49& 91.20& 41.65\\
        % \midrule
        $Vit$ & 
        84.42& 24.81& 88.08& 35.84& 83.80& 34.12& 89.31& 27.84\\ 
        % \midrule
         ResNet& 86.84& 24.26& 85.09& 26.72& 80.98& 32.37& 90.0& 34.09\\
        % \midrule
         Xception& 89.38& 39.62& 88.95& 40.15& 79.39& 23.0& 94.24& 40.08\\
        % \midrule
         MobileNet& 86.08& 28.32& 86.57& 35.29& 80.57& 28.97& 88.78& 34.37\\
        % \midrule
         DensNet& 87.59& 38.21& 87.95& 25.04& 81.69& 35.85& 91.66& 41.64\\
        % \midrule
         Siamese& 58.73& 12.84& 58.74& 12.85& 49.2& 11.6& 70.97& 12.22\\
        \midrule
        $HomNet$ &
        \vpara{91.64}& \vpara{52.91}& \vpara{95.73}& \vpara{66.30}& \vpara{91.64}& \vpara{52.65}& \vpara{98.25}& \vpara{66.32}\\ 
        \bottomrule
        \end{tabular}
    \end{threeparttable}
    }
    \label{tab:main_resualt}
\end{table}

\begin{table}[htbp]
    \centering
    \caption{Experimental results on Pki-3 with artificial structural abnormalities.}
    \resizebox{1.0\columnwidth}{!}{
    \setlength{\tabcolsep}{0.8pt}
    \begin{threeparttable}
        \begin{tabular}{c c c c c c c}
        \toprule
        Method& LR& OC-SVM& Deep-SVDD& AE& MLP-Mixer & Vit\\
        AUC& 60.82& 45.72& 48.51& 49.73& 50.22& 40.51\\
        F1&  53.79& 49.38& 37.59& 15.66& 43.53& 51.89\\
        \midrule
        Method& ResNet& Xception& MobileNet& DensNet & Siamese & \modelname\\
        AUC& 51.68& 48.88& 70.31& 50.58& 74.09& \vpara{75.71}\\
        F1& 38.28& 38.15& 58.97& 31.87& 59.95& \vpara{64.04}\\
        \bottomrule
        \end{tabular}
    \end{threeparttable}
    }
    \label{tab:pki-3}
\end{table}

\subsection{Performance Evaluation}
We present the performance of all methods in chromosomal structural abnormalities detection across four hospital datasets in \tableref{tab:main_resualt}. 
The results for $Hos\#4$ outperform those of other hospitals, while the outcomes for $Hos\#1$ and $Hos\#3$ are marginally less favorable. 
The varying performances across different hospitals highlight the inconsistency in the distribution of chromosome data, emphasizing the need to individually fine-tune the data from each hospital based on these experimental findings.

Overall, \modelname demonstrates superior performance across all four hospitals. 
On the one hand, the manual feature-based logistic regression method only exhibits limited capability in detecting chromosomal structural abnormalities. 
On the other hand, the traditional anomaly detection method OC-SVM performs poorly, facing challenges in detecting anomalies in high-dimensional data~\cite{zimek2012survey}, especially in the context of the $4\times512$ dimension of homologous chromosomes data.  
Similarly, the performance of the deep anomaly detection models Deep-SVDD and the AE-based model lags behind that of logistic regression, as they primarily capture characteristics of normal chromosomes during training, without effectively leveraging the information from structural abnormalities.
The Siamese method, utilizing contrastive learning, fails to yield optimal results as it does not incorporate a dedicated feature extraction backbone network tailored to chromosome characteristics.
In the case of the MLP-Mixer network, which use the strategy of dividing patches resembling chromosome regions, results in suboptimal performance across most experiments.
In contrast, \modelname leverages the homologous similarity of chromosomes for diagnosing structural abnormalities, while also effectively modeling the chromosome regions. 
\modelname also considers information from multiple pairs of homologous chromosomes from multiple cells, thereby enabling it to obtain more reliable diagnostic results. 
Therefore, it performs the best in both AUC-ROC and F1 score.

The experimental results for all baseline methods and \modelname on the public dataset Pik-3 with artificial structural abnormalities are presented in \tableref{tab:pki-3}. 
Once again, \modelname emerges as the top performer. 
Notably, in comparison to experiments conducted on datasets with real abnormalities, Siamese demonstrates improved performance on Pik-3 and almost rivals \modelname. 
This discrepancy may arise from a distribution shift between artificial and real structural abnormalities. 
This observation further emphasizes the importance of fine-tuning on datasets with real structural abnormalitis after pre-training with artificial abnormalities.

\begin{table}[htbp]
    \centering
    \caption{In-deep analysis on the difference of \modelname. The ``w/o Pair" indicates that only one chromosome is used, but the bag strategy using chromosome data from multiple cells is still used. The ``w/o Bag" indicates that the bag strategy is not used. ``align$\rightarrow$MLP" and ``align$\rightarrow$CNN" rows indicate that we replace the alignment part of \modelname with MLP and convolution neural networks. The last five rows changes the $\text{\#merge\ range}$ in the $CMSBlock$ with difference length.}
    \resizebox{1.0\columnwidth}{!}{
        \begin{tabular}{cc cc cc cc c}
        \toprule
        % first line
        \multirow{2}{*}{\diagbox{Variant}{Hospital}} & 
        \multicolumn{2}{ c}{$Hos\#1$}& 
        \multicolumn{2}{ c}{$Hos\#2$}& 
        \multicolumn{2}{ c}{$Hos\#3$}& 
        \multicolumn{2}{ c}{$Hos\#4$}\\
        % hos_1 for ZD3, hos_2 for ZD1, hos_3 for JH, hos_4 for NB
        % second line
        & AUC& F1
        & AUC& F1
        & AUC& F1
        & AUC& F1 \\
        \midrule
        w/o Pair & 
        % hos_1 for ZD3, hos_2 for ZD1, hos_3 for JH, hos_4 for NB
        % AUC   F1  |  AUC     F1 |   AUC   F1  |   AUC  F1
        78.13& 17.45& 79.42& 16.42& 75.31& 19.48& 91.63& 16.51\\
        \midrule
        \multirow{2}{*}{}
        align$\rightarrow$MLP & 
        % hos_1 for ZD3, hos_2 for ZD1, hos_3 for JH, hos_4 for NB
        % AUC   F1  |  AUC     F1 |   AUC   F1  |   AUC  F1
        87.82& 46.98& 86.19& 39.20& 71.40& 28.17& 97.32& 58.65\\
        align$\rightarrow$CNN &
        % hos_1 for ZD3, hos_2 for ZD1, hos_3 for JH, hos_4 for NB
        % AUC   F1  |  AUC     F1 |   AUC   F1  |   AUC  F1
        78.12& 17.42& 79.43& 16.38& 75.33& 19.82& 91.65& 16.45\\ 
        \midrule 
        w/o Bag & 
        % hos_1 for ZD3, hos_2 for ZD1, hos_3 for JH, hos_4 for NB
        % AUC   F1  |  AUC     F1 |   AUC   F1  |   AUC  F1
        88.91& 42.05& 88.13& 45.28& 88.23& 47.45& 93.91& 46.30\\
        \midrule
        \multirow{3}{*}{} 
        {$\text{range\#8}$} &
        % hos_1 for ZD3, hos_2 for ZD1, hos_3 for JH, hos_4 for NB
        % AUC   F1  |  AUC     F1 |   AUC   F1  |   AUC  F1
        90.41& 52.72& 94.03& 62.81& 91.52& 51.91& 98.19& \vpara{66.76}\\
        {$\text{range\#16}$} & 
        % hos_1 for ZD3, hos_2 for ZD1, hos_3 for JH, hos_4 for NB
        % AUC   F1  |  AUC     F1 |   AUC   F1  |   AUC  F1
        91.43& 46.12& 92.34& 58.85& 91.12& 50.78& 97.04& 58.00\\
        {$\text{range\#32}$} &
        % hos_1 for ZD3, hos_2 for ZD1, hos_3 for JH, hos_4 for NB
        % AUC   F1  |  AUC     F1 |   AUC   F1  |   AUC  F1
        \vpara{91.64}& \vpara{52.91}& \vpara{95.73}& \vpara{66.30}& \vpara{91.64}& \vpara{52.65}& \vpara{98.25}& 66.32\\ 
        {$\text{range\#48}$} &
        % hos_1 for ZD3, hos_2 for ZD1, hos_3 for JH, hos_4 for NB
        % AUC   F1  |  AUC     F1 |   AUC   F1  |   AUC  F1
        58.32& 13.56& 67.63& 17.82& 60.51& 13.73& 74.04& 15.07\\
        {$\text{range\#64}$} &
        % hos_1 for ZD3, hos_2 for ZD1, hos_3 for JH, hos_4 for NB
        % AUC   F1  |  AUC     F1 |   AUC   F1  |   AUC  F1
        50.61& 11.03& 54.64& 12.06& 59.60& 12.13& 49.62& 9.45\\
        \bottomrule
        \end{tabular}
    }
    \label{tab:ablation_result}
\end{table}

\subsection{Ablation Study} 
\modelname utilizes the homologous similarity between aligned homologous chromosomes (Pair strategy) to diagnose structural abnormalities.
Additionally, \modelname considers multiple pairs of homologous chromosome information (Bag strategy) to improve prediction reliability and reduce noise disturbance, rather than just a single pair of homologous chromosomes.
To investigate the impact of the Pair strategy, Bag strategy and other part of \modelname on the performance, we conduct ablation studies in this section.

\vpara{Effect of homologous similarity of homologous chromosomes (Pair strategy)}.
Inspired by domain knowledge indicating that homologous chromosomes with structural abnormalities differ from each other, we designed \modelname to diagnose structural abnormalities by comparing the differences between a pair of homologous chromosomes.
To investigate whether comparing the differences based on homologous similarity is effective, we detect structural abnormalities using data from just a single chromosome in a homologous pair while retaining the Bag strategy, which involves considering multiple chromosomes simultaneously.
According to the results shown in \tableref{tab:ablation_result}, without comparing homologous chromosomes, the model's performance decreased the most.
Therefore, comparing homologous chromosomes based on similarity is the most crucial aspect of the model for detecting structural abnormalities.

\vpara{Effect of alignment}. 
According to \tableref{tab:main_resualt}, the performance of methods based on supervised learning is considerably worse than that of \modelname.
One of the main reasons is that they do not align homologous chromosomes but directly extract features and then make judgments.
\modelname extracts the difference between regions of homologous chromosomes using an adaptive alignment based on attention mechanism.
We compare the model with replacements for attention using MLP and Convolutional Neural Networks (CNNs).
According to the result shown in \tableref{tab:ablation_result}~, in the absence of an aligned model, the performance degradation is large and MLP outperforms CNNs.

\vpara{Effect of multiple pairs (Bag strategy) of homologous chromosomes}. 
When human experts diagnose chromosomes with structural abnormalities, they observe information from multiple pairs of homologous chromosomes to make more reliable judgments. 
The proposed model \modelname also takes advantage of this idea by considering information from multiple pairs of homologous chromosomes simultaneously. 
To intuitively demonstrate the effectiveness of multiple pairs, we use information from only one pair of homologous chromosomes.
According to the result shown in \tableref{tab:ablation_result}~, it is important to utilize information from multiple pairs of homologous chromosomes simultaneously.

\vpara{The appropriate $\text{\#merge\ range}$}. 
As mentioned in \secref{text:Methodology}, we divide $H_{i, j}^{mg}$ into multiple subsequences, each representing a region of the chromosome with a length of $\text{\#merge\ range}$.
In order to explore the most appropriate $\text{\#merge\ range}$, we conduct experiments with various ranges . 
In \modelname, we set the $\text{\#merge\ range}$ to 32 . 
According to the result shown in \tableref{tab:ablation_result}~, When $\text{\#merge\ range}$ is too small, such as 8 or 16, it may not be sufficient to cover a region, resulting in a slight degradation in \modelname's performance.
For $Hos\#4$, the F1 score of the \modelname improves by 0.44\% when $\text{\#merge\ range}$ is set as 8. 
However, it should be noted that the $\text{\#merge\ range}$ is one quarter of the original \modelname, so the number of subsequences will increase fourfold, requiring more training time and GPU memory space.
When $\text{\#merge\ range}$ is set too large, such as 48 or 64, they may cover more than one region, causing information loss or compression, and thus leading to a decrease in performance.
Therefore, when $\text{\#merge\ range}$ is too large, the performance decreases the most. 
Overall, an $\text{\#merge\ range}$ of 32 for the \modelname is the most appropriate setting.

\subsection{Running Time Comparison} 
We evaluated the running time of HomNet in comparison with other models by testing the completion of 100 batches and measuring the convergence time on the pre-training dataset.  The following results, presented in seconds, are obtained:

\begin{table}[htbp]
\centering
\label{tab:transposed_performance_correct}
\begin{tabular}{|l|c|c|}
\hline
\textbf{Method} & \textbf{100 batches} & \textbf{Convergence} \\ \hline
\textbf{LR} & 0.08 & 3543 \\ \hline
\textbf{OC-SVM} & 41.16 & 23812 \\ \hline
\textbf{Deep-SVDD} & 11.72 & 5067 \\ \hline
\textbf{AE} & 16.51 & 11398 \\ \hline
\textbf{MLP-Mixer} & 1.62 & 1549 \\ \hline
\textbf{Vit} & 2.45 & 3832 \\ \hline
\textbf{Xception} & 9.33 & 9772 \\ \hline
\textbf{MobileNet} & 1.80 & 5948 \\ \hline
\textbf{DenseNet} & 13.49 & 10998 \\ \hline
\textbf{Siamese} & 18.18 & 10229 \\ \hline
\textbf{HomNet} & 5.39 & 2468 \\ \hline
\end{tabular}
\caption{Running Time Comparison}
\end{table}

While HomNet is not the fastest, it requires a comparatively lesser amount of time. And it is more time-efficient than Siamese, another model for diagnosing chromosomal structural abnormalities.
For clinical diagnostic applications (as presented in Section 5), HomNet is capable of diagnosing a patient in under 5 milliseconds, which comfortably meets the basic speed requirements.
\section{Application}
\begin{figure}[ht]
    \centering
    \includegraphics[width=\linewidth]{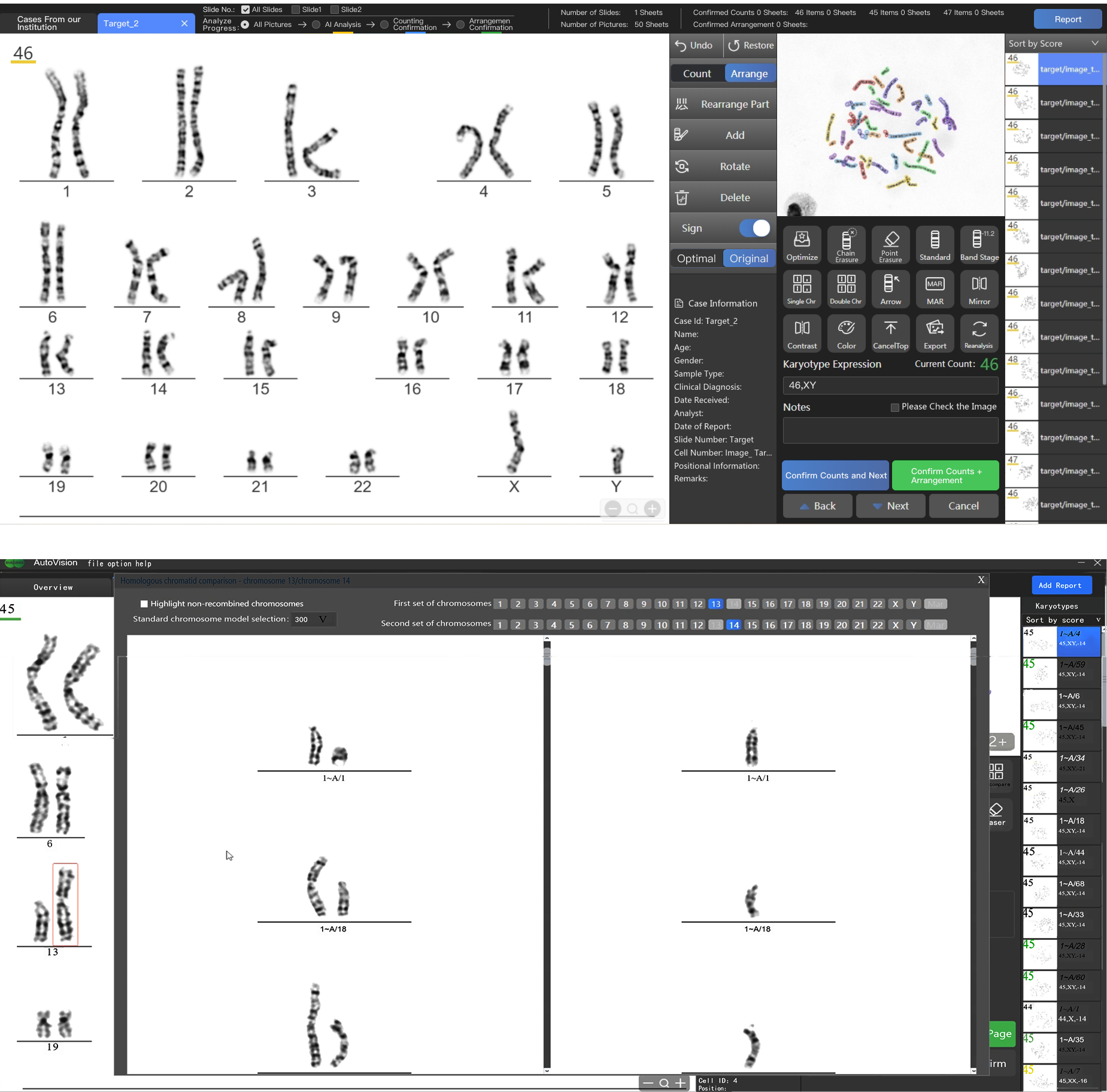}
    \caption{AutoVision, equipped with \modelname, is an advanced intelligent chromosome karyotype analysis system, which registered in NMPA (National Medical Products Administration)\protect\footnotemark and obtained CE mark\protect\footnotemark.}% The green boxes implying the }
    \label{fig:autovision} 
    \Description{Autovision equipped with \modelname.}
\end{figure}
\footnotetext{{NMPA is a national bureau responsible for drug supervision under the State Council of China.}}
\footnotetext{The CE mark on a product indicates that the manufacturer or importer of that product affirms its compliance with the relevant EU legislation.}

For practical clinical diagnostic applications, \modelname integrates an end-to-end pipeline with AutoVision~\cite{song2021chromosome}, an AI analysis system for karyotypes developed by Hangzhou Diagens Biotechnology Co., Ltd., China.
To begin with, physicians upload their patients' karyotypes to AutoVision, which has been enhanced with the integration of HomNet. 
Subsequently, the enhanced AutoVision, now incorporating HomNet, optimizes and organizes the karyotypes, yielding chromosome diagrams and band level identifications within an impressive 3 seconds, as depicted in Figure ~\figref{fig:autovision}.
The chromosome diagrams are then converted into sequence data, as detailed in ~\secref{text:Dataset}. 
This sequence data, along with the chromosome type and band levels, serves as input for \modelname, which diagnoses the presence of structural abnormalities. 
The final diagnosis results are directly output to the physicians. 
Notably, \modelname can diagnose chromosomal structural abnormalities using data from a single patient, encompassing 115 pairs of homologous chromosomes from five cells, in less than 5 milliseconds.

AutoVision, integrated with \modelname, has undergone multicenter clinical trials in three Grade A tertiary hospitals in East, Central, and West China. 
Presently, 1518 cases have been enrolled, with 499 sample analyses completed.
Preliminary results demonstrate that AutoVision exhibits 75\% sensitivity\footnote{Sensitivity (true positive rate) is the probability of a positive test result, conditioned on the individual truly being positive.}, 99\% specificity\footnote{Specificity (true negative rate) is the probability of a negative test result, conditioned on the individual truly being negative.}, and a diagnostic accuracy of 98.4\% in automated analysis, showcasing its precision and efficiency in clinical applications.
The product was honored with the first prize in technological achievements by the China Birth Defects Intervention Assistance Fund.

\section{Related Work}
For clarity, we first review methods related to anomaly detection, then introduce works about detecting chromosomes with structural abnormalities.

Anomaly detection is an important topic in diverse research fields and application domains~\cite{chai2022can, NEURIPS2022_8f1918f7, BrainNet}. 
Anomalies are also referred to as outliers , deviants or abnormalities~\cite{aggarwal2017introduction}.
Chromosomes with structural abnormalities that are different from normal chromosomes can also be called anomalies. 
 
Traditional anomaly detection methods include Principal Component Analysis (PCA) based methods~\cite{hoffmann2007kernel,shyu2003novel}, Nearest Neighbor (NN) based methods~\cite{knorr2000distance,breunig2000lof}, and the One-Class Support Vector Machine (OC-SVM)~\cite{scholkopf2001estimating} that learns a decision function for novelty detection, that is, classifying new data as similar or different one to the training set. 
These methods are unsupervised. 
Because in the anomaly detection setting, even if abnormalities are available, it is usually insufficient to make a supervised approach ffective~\cite{ruff2021unifying}.

As deep learning~\cite{lecun2015deep} has made substantial progress in dealing with many machine learning problems, some deep learning-based anomaly detection methods have also been proposed. 
A substantial amount of deep learning methods on anomaly detection choose to use autoencoder (AE) and its variants~\cite{principi2017acoustic,pawlowski2018unsupervised}. 
Deep Support Vector Data Description (Deep-SVDD)~\cite{ruff2018deep}, a variant of one-class neural network methods, trains a neural network and minimizes the volume of a hypersphere that maps the normal instances closely to the center of sphere.  
Besides, there are some other popular deep one-class classification based approaches~\cite{erfani2016high,sabokrou2018adversarially,oza2018one}. 
Method based on one deep generative models~\cite{schlegl2017unsupervised,schlegl2019f,zenati2018adversarially} and self-supervised methods~\cite{golan2018deep,tack2020csi} are also applied in anomaly detection. 

In recent years, some anomaly detection methods have been applied in the medical field~\cite{sato2018primitive,khan2018review}. 
These methods employed state-of-the-state techniques in anomaly detection and improved performance compared to traditional methods. 
But only a few works focus on chromosomal structural abnormality detection. 
~\cite{bechar2023highly} proposes to use the method of contrastive learning to detect chromosomal structural abnormalities using Siamese network.
Most of these methods do not take advantage of the homologous similarity between homologous chromosomes. 
\section{Conclusion}
In this paper, we study the problem of chromosomal structural abnormalities diagnosis. 
Inspired by the homologous similarity of chromosomes and the diagnostic process of doctors, we propose a novel model, \modelname, to model features of chromosomes and diagnose chromosomal structural abnormalities by homologous similarity. 

Central to our approach is the innovative utilization of multiple homologous chromosome pairs: a strategy akin to how skilled diagnosticians compare multiple diagrams for accuracy. \modelname simultaneously considers the information of multiple pairs of homologous chromosomes and detects structural abnormalities by capturing the differences between homologous chromosomes. 

Additionally, we artificially construct chromosomal samples with structural abnormalities according to different types of real-world structural abnormalities. 
And self-supervised learning has been adopted to enable \modelname to fully learn chromosomes with data of normal chromosomes and artificial structural abnormalities. Then, we fine-tune \modelname with data from different hospital to address the challenge of inconsistent distribution.

Experimental results show that \modelname achieved significantly better performance than other baselines on real-world dataset and public datset with artificial abnormalities. 
In addition, ablation studies further underscore the indispensability of each component within \modelname, highlighting the synergistic effect of its carefully crafted design.
% We hope that this work will bring insights into other anomaly detection with paired sample.
% \clearpage

\begin{acks}
This work is supported by National Natural Science Foundation of China (No. 62176233, No. 62441605) and SMP-IDATA Open Youth Fund.
\end{acks}

\clearpage
\bibliographystyle{ACM-Reference-Format}
\balance
\bibliography{cite}
\clearpage
\appendix
\section{Appendix}
\label{txt:appendix}
\subsection{Artificial abnormality}
According to the real world chromosomes with structural abnormalities in our datasets, we construct five types of artificial structural abnormality chromosome sequences:

\begin{itemize}[leftmargin=*]

\item{Deleted.} Part of the sampled chromosome sequences is deleted, which makes the length of the chromosome sequence shorter than before. The corresponding types of real world chromosome structural abnormality include deletion: a portion of the chromosome is missing or has been deleted ~\cite{2005Human}, and the donor chromosomes in insertion abnormalities: a portion of one chromosome has been deleted from its normal place and inserted into another chromosome ~\cite{banavali2013partial}. 

\item{Added with other fragments.} The sampled chromosome sequences are added with a non-self segment, which makes the length of the chromosome sequences longer than before. The corresponding types of real world chromosome structural abnormalities include the receptor chromosomes of insertion abnormality and some unbalanced translocations: a portion of one chromosome has been transferred to another chromosome ~\cite{milunsky2015genetic}. 

\item{Duplicated with self fragments.} The sampled chromosome sequences are added with a segment of its own, which also makes the chromosome sequences longer. The corresponding type of real world chromosome structural abnormality is duplication: a portion of the chromosome has been duplicated ~\cite{zhang2003evolution}. 

\item{Replaced.} One segment of a pair of sampled chromosome sequences, the left part and right part, is replaced by another, which doesn't change the length. If it is replaced by a segment of itself, the corresponding type of real world chromosome structural abnormality is inversion: A portion of the chromosome has broken off, turned upside down, and reattached ~\cite{painter1933new}. If it is replaced by a segment from another pair of sampled chromosome sequences, the corresponding type of real world chromosome structural abnormality is balanced translocations. 

\item{Robertsonian translocation.} The chromosome sampling sequence was symmetrical and repeated, which make the length doubled. The corresponding type of real world chromosome abnormality is robertsonian translocation: an entire chromosome has attached to another at the centromere ~\cite{hartwell2018genetics}.
\end{itemize}

\subsection{Datasets}
As mentioned in \secref{text:Dataset}, we use the dataset sampled from both healthy people and patients. 
% Appendix
We split the datasets according to the desensitized patient IDs. 
There is no overlap between the patients in the training set and test set. 
In the self-supervised learning stage, the split ratio of training set and test set is about 9: 1. 
And chromosome data with artificial structural abnormalities can avoid data imbalance. 
Therefore, the ratio of normal chromosomes bag to bag of chromosomes with artificial structural abnormalities is balanced. 
In the fine tuning stage, we split the dataset into training set and test set with a ratio of 1: 4.
The ratio of chromosome with and without structural abnormalities is less than 4\%.
Each type of chromosome data for each patient corresponds to multiple query tuples $(X, c, b)$. 
For each type of chromosome of each person, there are multiple chromosome diagrams from different cell. 
In the dataset, the average number of homologous chromosomes pair of each type of chromosome from one person is 6 for the healthy person dataset of $Hos\#1$, because some of the chromosome photos are incomplete due to occlusion and some other reasons.
And the average number is 13 for the patients datasets of $Hos\#1$, $Hos\#2$, $Hos\#3$ and $Hos\#4$. 
In our experiment, in order to cover as many patients as possible in the datasets, we let each bag, $X$, contain five pairs of homologous chromosome. 
In other words, $m$, the number of homologous pairs, is set as 5. 
Additionally, for the same type of chromosome data of the same patient, we randomly combined multiple bags.

\subsection{Experiment setting}
% Appendix
For feature engineering-based LR, we carefully extract features of one chromosome including mean, variance, on-peak and off-peak of the sampled sequence data, and features between the chromosomes in a homologous pair including DTW distance, the correlation coefficient and covariance. 
For anomaly detection methods, such as OC-SVM, Deep-SVDD and AE-based method, we use the data of normal chromosomes from healthy people as the training set. 
For the supervised-learning based method, we change the operation of dividing the image into multiple patches in the first step into the operation of dividing the sampled sequence into multiple subsequences or transfer the Conv2D into Conv1D to handle the sequential chromosome data.
For LR and OC-SVM, We used the sk-learn~\cite{scikit-learn} library for our experiments.
For Deep-SVDD and AE-based method, we use the PyOD~\cite{zhao2019pyod} libary for our experiments.
The codes of MobileNet, Xception, DensNet and ResNet refer to \url{https://github.com/weiaicunzai/pytorch-cifar100}.
The code of MLP-Mixer refers to \url{https://github.com/google-research/vision\_transformer}.
The code of Vit refers to \url{https://github.com/lucidrains/vit-pytorch}.
The code of Siamese refers to \url{https://github.com/MEABECHAR/ChromosomeSiameseAD}.
Besides, we use Adam~\cite{kingma2014adam} optimizer with a learning rate of 0.001 for self-supervised learning and that of 0.00001 for fine-tuning, and the batch size is set as 512. 
During the fine-tuning, we freeze the parameters of $CMSBlock$ and the first difference extraction layer in $HomBlock$, namely the first attention layer.

\end{document}